\documentclass[letterpaper, 10 pt, conference]{ieeeconf}  
\usepackage{cite}
\usepackage{multirow}
\usepackage{multicol}
\usepackage{adjustbox}

\usepackage{amsmath}
\usepackage{epsfig} 
\usepackage{booktabs} 
\usepackage{graphicx}
\usepackage{svg}
\usepackage{pifont}
\usepackage{float}
\usepackage[bookmarks=true]{hyperref}
\usepackage{graphics} 
\usepackage[font=small]{caption}
\graphicspath{{Images/}}
\usepackage{amssymb}
\usepackage{bbm}
\usepackage{listings}
\usepackage[most]{tcolorbox}
\usepackage{colortbl}
\usepackage{xcolor}
\definecolor{dropcolor}{RGB}{110, 135, 160}
\definecolor{tabfirst}{rgb}{0.7, 0.85, 0.95}
\definecolor{tabsecond}{rgb}{0.88, 0.95, 1.0}
\definecolor{tabthird}{rgb}{1, 1, 0.7}
\newcommand{\cfirst}{\cellcolor{tabfirst}\bfseries}
\newcommand{\csecond}{\cellcolor{tabsecond}}

\usepackage{mathptmx} 
\usepackage{times} 
\usepackage{marvosym}

\newtcolorbox{promptbox}[1]{
    colback=gray!5,          
    colframe=gray!50,        
    coltitle=black,          
    title={\textbf{#1}},     
    boxrule=0.5pt,          
    arc=1mm,                 
    left=1mm, right=1mm, top=1mm, bottom=1mm, 
    fontupper=\ttfamily\small, 
    breakable,              
    enhanced           
}

\IEEEoverridecommandlockouts                              

\title{\LARGE \bf
ThinkLocoManip: A Whole-Body Control Agentic Harness \\
for Legged Loco-Manipulation
}

\author{Nuo Chen\textsuperscript{1}, 
Shizeng Zhang\textsuperscript{1}, 
Yujie Yang\textsuperscript{1}, 
Jinghong Wang\textsuperscript{1}, 
Zhenyang Qu\textsuperscript{1}, 
Xiao-Ming Wu\textsuperscript{2}, 
Yunfei Guo\textsuperscript{3}, \\
Zhongwen Lu\textsuperscript{1}, 
Haojie Liu\textsuperscript{1}, 
Chunpeng Lu\textsuperscript{1}, 
Qiao Sun\textsuperscript{1}, 
Xueyi Zhang\textsuperscript{1}, 
Jiangwei Zhong\textsuperscript{1,\Letter}
\thanks{\textsuperscript{1}Robotics Lab, Lenovo CTO Organization, China. 
Email: {\tt\small \{chennuo5, zhangsz10, yangyj32, wangjh57, quzy4, luzw5, liuhj31, lucp1, sunqiao6, zhangxy41, zhongjw\}@lenovo.com}}
\thanks{\textsuperscript{2}S-lab, Nanyang Technological University, Singapore. 
Email: {\tt\small xiaoming.wu@ntu.edu.sg}}
\thanks{\textsuperscript{3}Lab of Robotics Oriented Intelligence, Fudan University, China.
Email: {\tt\small 25213050170@m.fudan.edu.cn}}
\thanks{\textsuperscript{\Letter}Corresponding author: Jiangwei Zhong {\tt\small zhongjw@lenovo.com}}
}

\begin{document}
\maketitle
\thispagestyle{empty}
\pagestyle{empty}

\begin{abstract}
In this paper, we propose ThinkLocoManip, an agentic harness for legged loco-manipulation designed to tackle complex, long-horizon tasks in large-scale environments. Existing approaches typically follow either direct action generation or a plan-then-act paradigm, both of which can struggle with the generalization and error-recovery capabilities required for extended task execution. In contrast, ThinkLocoManip adopts an interleaved reasoning-and-acting paradigm, allowing the agent to continuously reason about its current state, execute actions, observe their outcomes, and adapt its strategy accordingly. Building on this paradigm, our agentic harness orchestrates a suite of specialized tools for \textit{perception, navigation, manipulation, and locomotion}, enabling flexible composition of heterogeneous capabilities. Unlike existing robotic agentic harnesses, ThinkLocoManip explicitly incorporates locomotion as a tool domain for agent reasoning and execution. We further design a set of skills that explicitly account for both tool usage within each functional domain and coordination across domains, supporting effective execution of complex behaviors over long horizons.
To tackle task failure during long-horizon interactions, we develop keyframe thinking mechanism to preserve critical
execution context and support efficient error recovery.
Extensive real-world experiments demonstrate that ThinkLocoManip substantially outperforms representative methods based on both direct action generation and plan-then-act paradigms across challenging long-horizon legged loco-manipulation tasks. Notably, ThinkLocoManip achieves a success rate of 77\%, more than twice that of the strongest plan-then-act baseline (37.1\%). Our experiments further validate the effectiveness of interleaved reasoning and acting, tool orchestration, and coordinated skill execution.

\end{abstract}

\section{INTRODUCTION}

We are entering an era in which intelligent robots are expected to move beyond laboratories and operate autonomously in large-scale real-world environments, such as homes, supermarkets, factories, and hospitals. Performing labor-intensive and repetitive tasks, such as cleaning, organizing goods, and assisting humans, requires more than manipulating objects within a fixed workspace: robots must navigate through the environment, physically interact with it, and coordinate whole-body motion over extended task horizons. This capability, known as \textit{loco-manipulation}, tightly integrates locomotion and manipulation, enabling physical interactions beyond the reach of stationary robots. Robust and generalizable loco-manipulation is therefore a fundamental capability toward general-purpose embodied intelligence. In this work, we focus on \textit{legged loco-manipulation}, where legged mobility provides the flexibility and robustness required to traverse complex terrains.

Existing learning-based approaches to loco-manipulation broadly follow two paradigms: \textit{direct action generation} and \textit{plan-then-act}. Direct action generation typically learns an end-to-end policy through reinforcement learning or imitation learning, mapping observations directly to robot actions. While effective for reactive control, such policies often struggle to generalize to complex, long-horizon tasks in large-scale environments, largely due to the limited diversity of training data or simulation environments. In contrast, plan-then-act methods explicitly construct a high-level plan before execution, providing stronger task-level structure for complex tasks over extended horizons. However, they can be brittle when new observations invalidate the initial plan or when execution errors require recovery. Consequently, both paradigms remain limited in the generalization and error-recovery capabilities required for robust execution of complex, long-horizon tasks in large-scale environments.

\begin{figure}[t]
    \centering
    \includegraphics[width=0.49\textwidth]{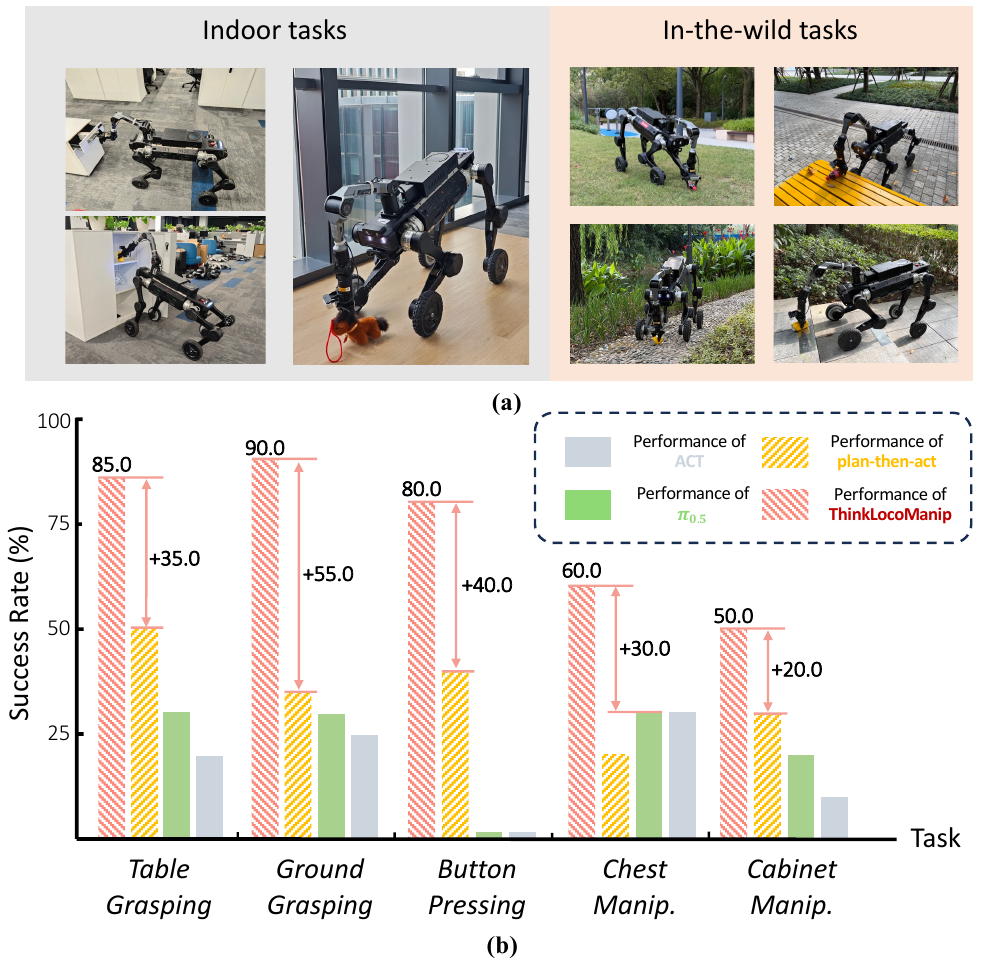}
    \caption{(a) An overview of our method for indoor and in-the-wild loco-manipulation tasks. (b) Performance comparison of long-horizon real-world tasks. }
    \label{fig:teaser}
    \vspace{-6mm}
\end{figure}

\begin{figure}[t]
    \centering
    \includegraphics[width=0.49\textwidth]{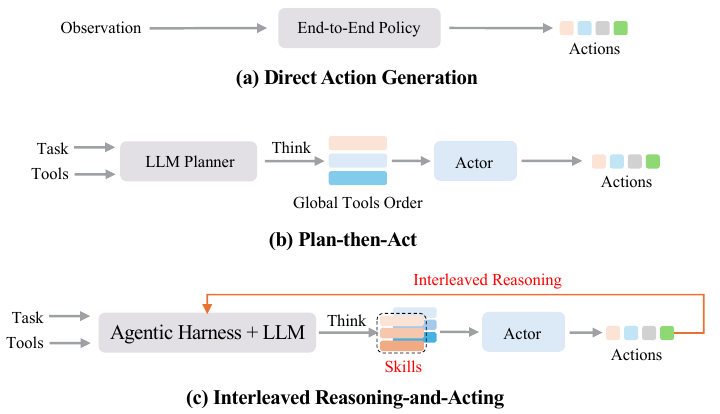}
    \caption{\textbf{The Comparison of different frameworks for legged loco-manipulation}. (a) direct action generation method uses an end-to-end policy to map observations directly to robot actions. (b) plan-then-act method constructs a tool invoking plans and then execute each tool. (c) Our approach design an agentic harness, enabling interleaved reasoning-and-acting.}
    \label{fig:comparison}
    \vspace{-7mm}
\end{figure}

Toward this end, we propose \textbf{ThinkLocoManip}, an agentic harness for legged loco-manipulation designed to tackle complex, long-horizon tasks in large-scale environments. In contrast to the two paradigms above, ThinkLocoManip adopts an interleaved reasoning-and-acting paradigm, allowing the robot to continuously reason about its current state, invoke appropriate tools, observe execution outcomes, and adapt its strategy accordingly, as seen in Fig. \ref{fig:comparison}. This paradigm enables stronger generalization across diverse environments and robust recovery from execution errors. To support such adaptive execution, we organize the toolset into four functional domains: \textit{perception, navigation, manipulation, and locomotion}. Unlike existing robotic agentic harnesses that primarily orchestrate perception, navigation, and manipulation capabilities, ThinkLocoManip further treats locomotion as a first-class tool domain, allowing the agent to explicitly reason about and invoke locomotion capabilities as part of task execution. Each domain comprises specialized tools that can be selected according to the task requirements and environmental conditions. For example, the robot can employ map-based navigation when a prior map is available, while target-based navigation allows it to actively approach and observe a target when such prior knowledge is unavailable. Building on this modular organization, we further design a set of skills that govern both tool usage within each domain and coordination across domains. In total, ThinkLocoManip integrates 11 tools and 5 skills, enabling flexible and coordinated execution of complex behaviors over long task horizons. We also develop Keyframe Thinking error-recovery mechanism that enables the agent to identify execution failures and adapt its behavior, thereby improving robustness over long-horizon task execution.

Extensive real-world experiments demonstrate that ThinkLocoManip substantially outperforms representative methods based on both direct action generation and plan-then-act paradigms across challenging long-horizon legged loco-manipulation tasks. In particular, ThinkLocoManip achieves an overall success rate of \textbf{77\%}, more than twice that of the strongest plan-then-act baseline at \textbf{37.1\%}, corresponding to an absolute improvement of nearly \textbf{40 percentage points}. Further analysis validates the effectiveness of interleaved reasoning and acting, tool orchestration, and coordinated skill execution, while the accompanying video demonstrates the versatility of ThinkLocoManip across diverse real-world tasks and environments.

\section{RELATED WORKS}

\subsection{Loco-Manipulation Policy}

Achieving general-purpose loco-manipulation remains a fundamental challenge for legged robots equipped with manipulators. Existing approaches can be broadly categorized into \textit{optimization-based control} and \textit{learning-based} methods. Traditional optimization-based approaches~\cite{Ferrolho2023roloma, Wolfslag2020optimisation, Sleiman2021unified} typically formulate loco-manipulation as a whole-body optimization problem and employ methods such as model predictive control (MPC) to jointly coordinate legged locomotion and arm manipulation. While providing precise and structured control, these methods can incur high computational costs and encounter infeasibility under hard constraints.

Learning-based approaches, in contrast, learn loco-manipulation capabilities from data and generally follow two paradigms. The first is \textit{direct action generation}, where reinforcement learning (RL) is commonly used to first train low-level whole-body controllers that maintain balance and stabilize the robot during motion~\cite{Fuzp2022deep, Portela2024learning, Houdy2025efficient, Pangp2025roboduet, Zhipy2025learning}. Building on these stabilized whole-body controllers, subsequent methods~\cite{Wangjl2025quadwbg, Zhangjz2024gamma, Liumh2024visual} further learn goal-conditioned loco-manipulation policies that coordinate locomotion and manipulation. To further improve manipulation capabilities, recent approaches~\cite{Liux2026mlm, Hah2024umi} incorporate imitation learning or vision-language-action models (VLAs) to learn manipulation policies from human demonstrations. Despite their effectiveness, these methods can struggle to generalize to diverse real-world environments and long-horizon tasks due to limited training-data diversity and the sim-to-real gap.

The second learning-based paradigm is \textit{plan-then-act}~\cite{Yokoyama2024asc, Qiurz2025learning, Qiurz2025wildlma, Wangkj2026odyssey}, which decomposes loco-manipulation into reusable primitive tools and employs large language models (LLMs) or vision-language models (VLMs) to compose them into high-level plans for complex tasks. Such modularization improves compositionality and enables more flexible task execution. However, because reasoning and execution are typically separated, these approaches can become brittle when observations invalidate the initial plan or execution errors occur, limiting their ability to recover during long-horizon interactions.

In contrast, ThinkLocoManip introduces an agentic harness for legged loco-manipulation based on an \textit{interleaved reasoning-and-acting} paradigm. Rather than committing to a fixed policy or a pre-generated plan, the agent continuously reasons about its current state, invokes appropriate tools, observes execution outcomes, and adapts its strategy accordingly. This design improves generalization to diverse environments while enabling robust recovery from execution errors and external interference.

\subsection{Agentic Harness for Robotics}
Leveraging LLMs’ semantic understanding and reasoning capabilities, agentic harnesses have emerged as a promising paradigm for robotic manipulation~\cite{Liry2026roboclaw, Huangjb2026roboharness}. Early works~\cite{Qiurz2025wildlma, liangj2023code, huanghx2024copa, Yangy2025maniagent} leverage LLMs to decompose high-level instructions into sequential sub-tasks and synthesize Python programs that invoke predefined action primitives. However, these methods typically follow an open-loop plan-then-act paradigm, in which the LLM generates the complete executable program upfront based on the task instruction. Consequently, they cannot adapt to intermediate execution outcomes, making them vulnerable to execution errors and environmental uncertainties.

\begin{figure*}[t]
    \centering
    \includegraphics[width=0.99\textwidth]{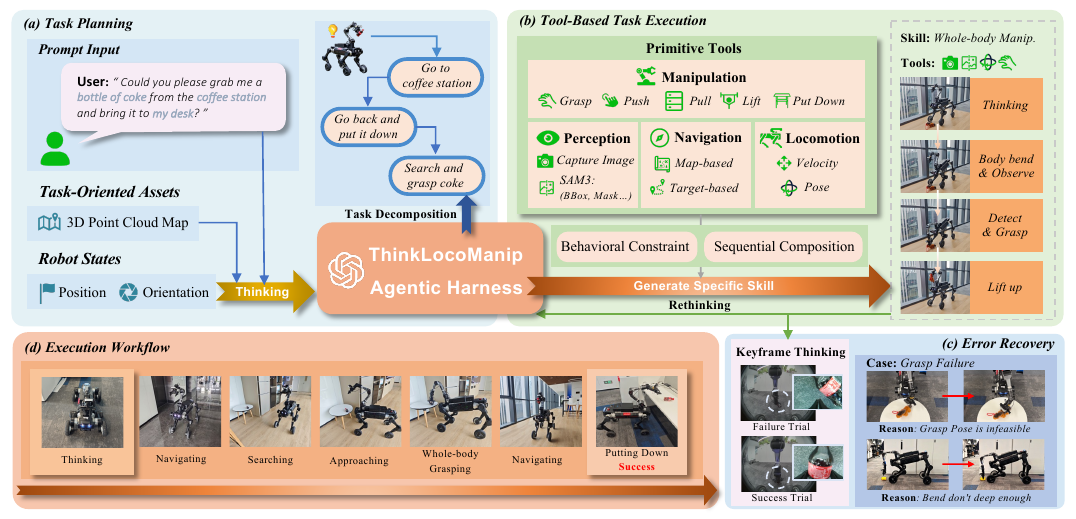}
    
    \vspace{-2mm}
    \caption{\textbf{The overall framework of our method}. (a) The agent performs task planning to understand task instruction and decompose it into several sub-tasks. (b) ThinkLocoManip orchestrates a suite of \textbf{Primitive Tools} for perception, navigation, manipulation, and locomotion. We also design \textbf{ThinkLocoManip-Skill} that explicitly accounts for both tool usage within each functional domain and coordination across domains. (c) Keyframe Thinking mechanism enables error recovery from task failure. (d) The whole execution workflow of our method.}
    \label{fig:framework}
    \vspace{-5mm}
\end{figure*}

To improve the robustness and generalizability of robotic task execution, subsequent methods~\cite{Huangwl2022inner, Huangwl2024rekep, Yangzj2025agentic, Zhangyx2026harnessvla, Heyx2026closed-loop} incorporate agentic harnesses to support task decomposition, closed-loop interaction, execution monitoring, and error recovery. Despite their success in tabletop manipulation, extending these approaches to legged loco-manipulation remains challenging, as long-horizon tasks in large-scale and unstructured environments require effective coordination between locomotion and manipulation.

In this work, we introduce \textbf{ThinkLocoManip}, an agentic harness for legged loco-manipulation that equips the agent with specialized tools for \textit{perception, navigation, manipulation, and locomotion}. Unlike existing robotic agentic harnesses, ThinkLocoManip explicitly incorporates \textit{locomotion} as a first-class tool domain, allowing the agent to reason about and invoke locomotion capabilities alongside other tools during task execution. By coordinating these heterogeneous capabilities, ThinkLocoManip enables robust long-horizon task execution in complex environments.

\section{METHOD}
ThinkLocoManip comprises three hierarchical components for tackling long-horizon legged loco-manipulation: \textit{primitive tools}, \textit{ThinkLocoManip-Skill}, and \textit{Keyframe Thinking}. We first provide a formal definition of the legged long-horizon loco-manipulation task in Sec. III-A. In Sec. III-B, we introduce a suite of primitive tools specialized for long-horizon loco-manipulation, covering \textit{perception, navigation, manipulation, and locomotion}. Sec. III-C presents \textbf{ThinkLocoManip-Skill}, which defines both the workflow for composing primitive tools within each skill and the behavioral constraints for coordinating different skills. Finally, in Sec. III-D, we introduce \textbf{Keyframe Thinking}, an error-recovery mechanism that enables the agent to identify execution failures and adapt its behavior, thereby improving robustness over long-horizon task execution.

\subsection{Task Definition} 

Given a natural-language task instruction $\ell$, long-horizon legged loco-manipulation requires an agent to autonomously coordinate locomotion and manipulation over an extended horizon to accomplish the specified task. We formulate this process as closed-loop decision making, where the agent produces an action decision $\mathbf{u}_t$ conditioned on the task instruction and its execution history:

\vspace{-0.3cm}
\begin{equation}
    \mathbf{u}_t = \pi(\ell, \mathbf{o}_{1:t}, \mathbf{u}_{1:t-1}),
\end{equation}
where $\mathbf{o}_t$ denotes the observation at time step $t$, and $\mathbf{u}_t$ denotes the agent-level decision that ultimately determines the locomotion and manipulator behaviors.

As illustrated in Fig.~\ref{fig:framework}, the agent first decomposes the task instruction into an ordered sequence of sub-tasks. To complete each sub-task, the agent first navigates the robot to the corresponding task location and then executes the required loco-manipulation behavior by coordinating the robot's body posture and manipulator end-effector pose. Since long-horizon execution can be disrupted by external interference or execution errors, the agent must continuously incorporate new observations, identify task failures, and replan from the current state to continue execution. The following sections detail the design of our \textbf{primitive tools}, \textbf{ThinkLocoManip-Skill}, and \textbf{Keyframe Thinking} mechanism.

\begin{table}[t]
  \centering
  \caption{\textbf{Library of Primitive Tools.}}
  \vspace{-2mm}
  \label{tab:primitives}
  \resizebox{\linewidth}{!}{
  \begin{tabular}{@{}l l@{}}
    \toprule
    \textbf{Primitive Tool} & \textbf{Function} \\
    \midrule
    \multicolumn{2}{@{}l}{\textit{\textbf{Perception}}} \\
    \texttt{Capture Image} & Take RGB/Depth Image using camera. \\
    \texttt{Prompt Detection} & Detect the object matching the prompt. \\

    \midrule
    \multicolumn{2}{@{}l}{\textit{\textbf{Navigation}}} \\
    \texttt{Map-based Nav} & Navigate to waypoint based on 3D Map. \\
    \texttt{Target-based Nav} & Navigate to $(x,y,z)$ in the robot's base frame. \\
    \midrule
    \multicolumn{2}{@{}l}{\textit{\textbf{Manipulation}}} \\
    \texttt{Grasp} & Grasp the selected object. \\
    \texttt{Lift Up} & Lift up the gripper ($+\Delta z$). \\
    \texttt{Put Down} & Put down the object, release the gripper ($-\Delta z$). \\
    \texttt{Push} & Close the gripper and move it ($+\Delta x, \pm\Delta y$). \\
    \texttt{Pull Back} & Close the gripper and move it back ($-\Delta x$). \\
    \midrule
    \multicolumn{2}{@{}l}{\textit{\textbf{Locomotion}}} \\
    \texttt{Velocity Control} & Move the robot forward, backward, rotate. \\
    \texttt{Pose Control} & control the robot to adjust its body posture. \\
    \bottomrule
  \end{tabular}
  }
  \vspace{-0.5cm}
\end{table}

\subsection{Library of Loco-Manipulaton Primitive Tools}

We decompose the capabilities required for the robot to execute long-horizon loco-manipulation tasks into four categories of primitive tools: \textit{perception}, \textit{navigation}, \textit{manipulation}, and \textit{locomotion}.
Each category is further divided into more specific tools according to their respective functionalities.
We provide a brief and specific introduction to each primitive tool in this section.

\noindent\textbf{Perceptive Tools for Visual Understanding.} To enhance the legged robot's environmental perception and understanding, we integrate multi-view cameras and SAM3\cite{Nicolas2025sam3} semantic segmentation model. 
\textit{Capture Image:} We mount cameras on both the robot's head and the end of its manipulator, allowing the agent to flexibly choose between them based on its needs. 
When searching for an object, the agent uses the head camera to capture RGB image. 
When grasping the object at close range, it switches to the end-effector camera to obtain more precise RGB and depth image of the object.  
\textit{Prompt Detection:} After acquiring the RGB image, the agent invokes the SAM3 model with a text prompt to obtain the object's bounding box (bbox), pixel-wise segmentation mask, and 3D position in the robot's base frame.

\noindent\textbf{Navigational Tools for Localization and Path Planning.}
The robot needs to autonomously navigate through the environment to reach the target position of each sub-task. We design two approaches to enable robot navigation: a map-based way and a target-based way.
\textit{Map-based Nav:} Based on an pre-built LiDAR-based robocentric 3D occupancy grid map\cite{Renyf2024rogmap}, the robot can quickly localize itself\cite{Xuw2022fastlio2} and annotate waypoints in the map. 
The agent then matches the location information in the task description to the corresponding waypoint names and finally reaches the task destination through path planning\cite{Zhoux2021egoplanner}.
\textit{Target-based Nav:} Unlike map-based nav, target-based nav does not need pre-built map to plan the path. It directly computes the relative angle and distance between the robot base and the target pose, and then plans a straight-line trajectory toward the target. 
We design a two-stage approaching method, in which the robot first adjusts its body orientation toward the target object and then moves straight toward it.

\noindent\textbf{Manipulative Tools for Diverse Physical Interaction.}
After reaching the task location, the agent selects and combines appropriate manipulation tools from the primitive tool set according to the specific sub-task requirements.
For grasping task, we employ EconomicGrasp\cite{Wuxm2024economic}, a lightweight and high-performance 6-dof grasp pose detection method\cite{Chenn2025motiongrasp, Wangal2025task}, to generate a suitable grasp pose for the selected object.
To provide the robot with more diverse physical interaction capabilities, we further design a set of manipulation tools, including lifting up, putting down, pushing and pulling, enabling the agent to perform more complex tasks such as opening a chest or cabinet.
However, it is still hard for robot to accomplish challenging loco-manipulation purely depending on manipulative tools without locomotion tools. 

\noindent\textbf{Locomotion Tools for Whole-Body Control.}
Unlike existing robotic agentic harnesses that primarily orchestrate perception, navigation, and manipulation capabilities, we explicitly expose \textit{locomotion} as a first-class tool domain, allowing the agent to reason about and actively control the robot's whole-body motion during task execution.
To enable stable movement and task-dependent body adjustment, we provide locomotion tools for both body velocity and pose control.
\textit{Velocity Control:}
We adopt HIM~\cite{Longjf2024himloco} to train a robust reinforcement learning policy for locomotion velocity control.
The agent invokes this tool by specifying the desired planar velocity and yaw rate, which are tracked by the low-level locomotion policy while maintaining stable legged motion.
This allows the agent to continuously adjust the robot's position and orientation, such as approaching a manipulation target, aligning the body with the target, or repositioning the robot when the current configuration is unsuitable for subsequent manipulation.
\textit{Pose Control:} 
We develop a matrix-based formulation to compute the leg joint angles required to achieve a desired body posture.
To adjust the body posture while maintaining stable foot contacts, we keep the foot positions fixed in the reference frame. 
For the $i$-th leg, the desired foot pose in its local frame is computed as
\begin{equation}
{}^{L_i}T_{F_i}^{d}
=
\left({}^{W}T_{B}^{d}{}^{B}T_{L_i}\right)^{-1}
{}^{W}T_{F_i}^{0},
\end{equation}
where ${}^{W}T_{B}^{d}$ is the desired body pose, ${}^{B}T_{L_i}$ is the fixed transformation from the body frame to the $i$-th leg frame, and ${}^{W}T_{F_i}^{0}$ is the initial foot pose in the reference frame.

The desired joint angles are then obtained through inverse kinematics:
\begin{equation}
\mathbf{q}_{i}^{d}
=
\mathrm{IK}_{i}\left({}^{L_i}\mathbf{p}_{F_i}^{d}\right),
\end{equation}
where ${}^{L_i}\mathbf{p}_{F_i}^{d}$ is the translation component of ${}^{L_i}T_{F_i}^{d}$, representing the desired foot position in the $i$-th leg frame, and $\mathbf{q}_{i}^{d}$ denotes the corresponding target joint angles. Finally, the computed target joint positions are directly sent to the joint motors for execution.

These primitive tools enable the agent to perceive the environment, navigate to target locations, and coordinate manipulation with body motion and posture adjustment. By selecting and composing appropriate tools according to sub-task requirements, the agent can execute diverse long-horizon loco-manipulation tasks.

\subsection{ThinkLocoManip-Skill for Long-Horizon Tasks}

Although the agent is given so many kinds of primitive tools, it is still challenging to pick suitable tools and compose them together in a rational order. The agent should be capable of decomposing a task into a sequence of sub-tasks, coordinating whole-body posture throughout long-horizon loco-manipulation, and recovering promptly from execution failures. Toward this end, we introduce \textbf{ThinkLocoManip-Skill}, defines both the workflow for composing primitive tools within each skill and the behavioral constraints for coordinating different skills.

The description specifies the functionality of each skill briefly, enabling the agent to efficiently match appropriate skills to corresponding sub-tasks.
To ensure that each skill is invoked under appropriate conditions, we define the pre-rule of each skill to specify when it should be triggered.
Then the agent follows the predefined workflow of skills and executes each tool sequentially.
To ensure safe and stable execution throughout the entire skill process, we set five basic skills for long-horizon loco-manipulation: \textit{Sub-task Planning} is responsible for decomposing the task instruction into a sequence of sub-tasks. After each sub-task is completed, the agent verifies whether the execution was successful. If a failure is detected, the agent replans the execution order of the remaining sub-tasks. \textit{Path Navigation} guides the robot to the target position by selecting appropriate navigation tools based on the requirements of each sub-task. For indoor navigation, the agent invokes map-based navigation to reach a pre-labeled waypoint. For sub-tasks that require approaching a target in either indoor or in-the-wild environments, the agent first detects the target's spatial coordinates and then invokes target-based navigation to approach it. We set the distance between the robot head and the target object position to 10 cm. \textit{Object Searching} handles cases in which the target object is not visible given the robot’s current orientation. It instructs the agent to invoke perception tools to detect the target object. If the target is not detected, the agent commands the robot to rotate \(45^\circ\) clockwise until the object is detected or the search attmpts reaches a upper limit.or the maximum number of search attempts is reached. \textit{Error Recovery} enables the agent to analyze the causes of execution failures based on task execution logs and replan sub-tasks accordingly. We design Keyframe Thinking mechanism to efficiently record important task execution logs, helping the agent identify execution failures and adapt its behavior. The detail is shown in Section III-D.
\textit{Whole-body Manipulation} executes manipulation tools of each sub-task while adjusting the robot's body pose by invoking locomotive tools to maintain balance and facilitate task execution. We formulate a equation that automatically computes the desired body pose from the target object's position expressed in the robot's body frame: 

\vspace{-5mm}
\begin{equation}
\theta = \operatorname{atan2}\left(z, \sqrt{x^2+y^2}\right),
\end{equation}

where $(x,y,z)$ denotes the target object's position in the body frame, and $\theta$ is the angle between the robot and the object. The agent then invokes the pose control tool with $\theta$ as the target angle to orient the robot's body toward the target object. We present a simplified skill template along with a description of its contents:

\begin{promptbox}{Skill: Whole-body Manipulation}
\textbf{Description:} Actively adjusts the robot’s body posture and invokes manipulation tools to complete the corresponding sub-tasks.

\textbf{Pre-Rule:} Only invoked after the object searching sub-task or path navigation sub-task is completed; Object's name and 3D position must be provided;

\textbf{Workflow:} For \textbf{Grasping} task, compute the corresponding robot pose based on the object’s spatial coordinates in the robot frame then execute grasp tool; For \textbf{Opening} task, execute pull action first, and then move the robot back a small distance;

\textbf{Constraints:} Not allowed to invoke velocity control tool except for \textbf{Chest/Cabinet Opening} task.
\end{promptbox}

These five skills organize primitive tools into structured workflows, with explicit invocation conditions and behavioral constraints guiding skill coordination. ThinkLocoManip-Skill thus enables the agent to manage sub-task execution, coordinate body posture and manipulation, and recover from execution failures during long-horizon loco-manipulation.

\subsection{Keyframe Thinking for Error Recovery}

During long-horizon loco-manipulation, execution failures can invalidate the planned sub-task sequence and interrupt task progress. To address this challenge, we introduce Keyframe Thinking, an error-recovery mechanism that enables the agent to analyze task execution logs, identify potential causes of failure, and replan the sub-task sequence accordingly.
We define a keyframe as a time instant at which the robot makes physical contact with an object during loco-manipulation task execution.
Then the agent logs the RGB image, robot states, current sub-task and primitive tool's name into a file.
Rather than recording every frame during task execution, we log information only at keyframes, preserving essential task context while reducing the agent’s reasoning time.
After each sub-task execution, the resulting state is recorded and provided to the agent alongside the keyframe information to determine whether the task has been successfully completed.

\section{EXPERIMENTS}
\begin{figure*}[t]
    \centering
    \includegraphics[width=0.98\textwidth]{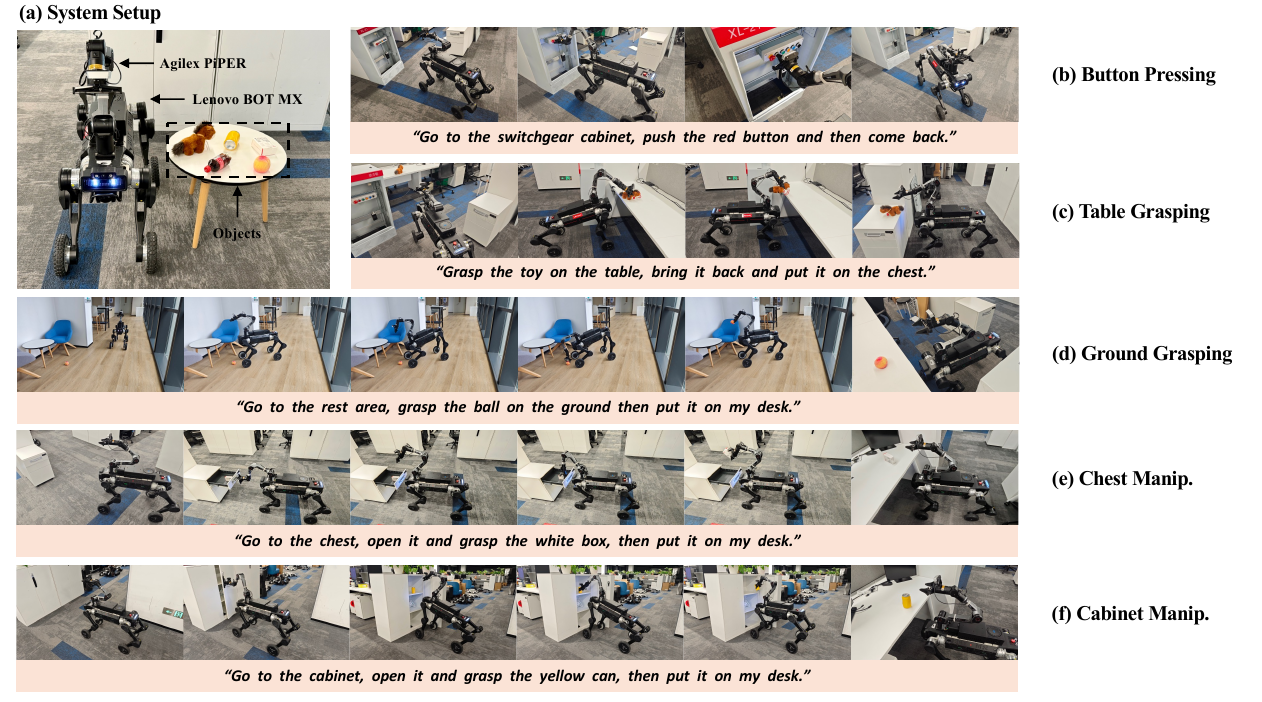}
    \vspace{-2mm}
    \caption{\textbf{The system setup and execution process of five real-world indoor tasks}. (a) We demonstrate the robot and objects used in the experiment. (b)-(f) We conduct experiments on five indoor tasks, illustrating the complete execution process from navigation to task completion and demonstrating the effectiveness of our agentic harness framework.}
    \label{fig:indoor}
    \vspace{-3mm}
\end{figure*}

\begin{table*}[t]
\centering
\caption{
Performance comparison on real-world long-horizon loco-manipulation tasks.
We color the \colorbox{tabfirst}{\textbf{best}} and
\colorbox{tabsecond}{second best} results.
The improvement in average success rate over the second-best method is shown in parentheses.
}
\vspace{-2mm}
\label{tab:main}

\setlength{\tabcolsep}{4pt}

\begin{tabular}{l|cc cc cc cc cc|cc}
\toprule
\multirow{2}{*}{\textbf{Method}} &
\multicolumn{2}{c}{\textbf{Table Grasping}} &
\multicolumn{2}{c}{\textbf{Ground Grasping}} &
\multicolumn{2}{c}{\textbf{Button Pressing}} &
\multicolumn{2}{c}{\textbf{Chest Manip.}} &
\multicolumn{2}{c}{\textbf{Cabinet Manip.}} &
\multicolumn{2}{c}{\textbf{Avg.}} \\
\cmidrule(lr){2-3}
\cmidrule(lr){4-5}
\cmidrule(lr){6-7}
\cmidrule(lr){8-9}
\cmidrule(lr){10-11}
\cmidrule(lr){12-13}

& Succ & TaskC (\%)
& Succ & TaskC (\%)
& Succ & TaskC (\%)
& Succ & TaskC (\%)
& Succ & TaskC (\%)
& SuccR (\%) & TaskC (\%) \\
\midrule

ACT
& 4/20 & 40.0
& 5/20 & 43.8
& 0/10 & 33.3
& \csecond{3/10} & 44.0
& 1/10 & 28.0
& 18.6 & 37.8 \\

\addlinespace[2pt]

$\pi_{0.5}$
& 6/20 & 47.5
& 6/20 & 47.5
& 0/10 & 33.3
& \csecond{3/10} & 47.5
& 2/10 & 40.0
& 24.3 & 43.2 \\

\addlinespace[2pt]

plan-then-act
& \csecond{10/20} & \csecond{62.5}
& \csecond{7/20} & \csecond{51.3}
& \csecond{4/10} & \csecond{60.0}
& 2/10 & \csecond{48.0}
& \csecond{3/10} & \csecond{56.0}
& \csecond{37.1} & \csecond{55.6} \\

\midrule

\textbf{Ours}
& \cfirst{17/20} & \cfirst{88.8}
& \cfirst{18/20} & \cfirst{92.5}
& \cfirst{8/10} & \cfirst{86.7}
& \cfirst{6/10} & \cfirst{72.0}
& \cfirst{5/10} & \cfirst{66.0}
& \cfirst{77.0} & \cfirst{81.2} \\

\bottomrule
\end{tabular}

\vspace{-5mm}
\end{table*}

\subsection{Experimental Setup}
In the main setup, we conduct five diverse \textbf{long-horizon loco-manipulation tasks in real-world indoor environments} to systematically evaluate our method. These tasks require the robot to autonomously execute multiple stages of navigation, whole-body adjustment, and manipulation over extended task horizons. All experiments are performed on the Lenovo BOT MX, a wheeled quadruped robot equipped with an Agilex PiPER 6-axis robotic arm mounted at the front of the robot, as shown in Fig.~\ref{fig:indoor}. We use GPT-5.6 Sol as the LLM backbone of our agent. At the beginning of each task, the agent is provided only with a natural-language task instruction and must autonomously control the robot throughout the entire task execution. According to task complexity, we categorize the five tasks into three simple tasks and two challenging tasks.

\paragraph{Simple Tasks}
(1) \textit{Table Grasping}: The robot is required to navigate to a table, grasp a specified object, and place it at a designated position. We evaluate the task using tables of two different heights, requiring the agent to adapt the robot's whole-body configuration to different manipulation heights.
(2) \textit{Ground Grasping}: The target object is placed on the ground. The agent must first locate and approach the object, and then adjust the robot's body posture to reach a feasible configuration for grasping.
(3) \textit{Button Pressing}: The robot is required to navigate to a switchgear cabinet, identify and press the button of the specified color, and subsequently navigate back.

\paragraph{Challenging Tasks}
(1) \textit{Chest Manipulation}: The robot is required to navigate to the chest, grasp its handle and pull it open, retrieve the target object from inside, and finally place the object on the table. This task requires the agent to coordinate multiple sequential navigation and manipulation behaviors over a longer execution horizon.
(2) \textit{Cabinet Manipulation}: Similar to \textit{Chest Manipulation}, the robot must navigate to the cabinet, grasp its handle and pull the cabinet open, and then retrieve the target object from inside. The task requires coordinated whole-body adjustment and manipulation across multiple stages.

We compare three method as baselines: (1) ACT\cite{Zhaozt2023act}, a visormotor policy for robotics manipulation. (2)$\pi_{0.5}$\cite{pi0.5}, a vision-language-action(VLA) model trained on a very large dataset. (3) plan-then-act: the LLMs constructs a high-level plan according to task and execute each tool.
ACT and $\pi_{0.5}$ represent learning-based methods that directly generate robot actions, while plan-then-act represents an LLM-based planning approach that directly operates on the provided primitive tools.
For each task, we report the number of successful trials (\textbf{Succ}), the success rate (\textbf{SuccR}), and the completion rate (\textbf{TaskC.}) of individual sub-tasks to provide a detailed evaluation of task execution performance.

\vspace{-3mm}
\subsection{Indoor Long-Horizon Task}
We set five indoor tasks to assess the performance of the method. 
Table Grasping and Ground Grasping tasks are tested for 20 times, the rest of them are tested for 10 times.
For each single manipulation sub-task, we collect 50 demonstration trajectories for ACT training and $\pi_{0.5}$ finetuning.
The results are shown in Table \ref{tab:main}. 
We can see that learning-based methods get poor performance because the robot's execution pose may deviate from the poses observed during data collection due to limited navigation accuracy, leading to distribution shifts in visual observations and object-relative configurations. Such deviations can significantly degrade the performance of learning-based policies that are sensitive to the initial robot pose, especially in the \textit{Button Pressing} task which requires high-precision manipulation.
Overall, our method achieves great performance compared with other baseline methods. We suppose the reasons are that: (1) a diverse and well-designed set of primitive tools. (2) ThinkLocoManip-Skill efficiently facilitates the robot’s execution of long-horizon loco-manipulation tasks. (3) Keyframe Thinking mechanism effectively enables the robot to recover from execution failures.

\vspace{-0.2cm}
\subsection{Ablation Study}

\begin{table}[t]
\centering
\caption{Ablation Study.}
\vspace{-2mm}
\label{tab:ablation}
\setlength{\tabcolsep}{3pt}
\begin{tabular}{l|c c c|c}
    \toprule
    \multirow{2}{*}{\textbf{{Ablated Module}}} &
    \textbf{Table} &
    \textbf{Ground} &
    \textbf{Chest} & 
    \multirow{2}{*}{\textbf{SuccR} (\%)} \\
    &
    \textbf{Grasping} &
    \textbf{Grasping} &
    \textbf{Manip.} & 
    \\    
    \midrule

{w/o Whole-body Control} &  12/20 &  0/20 &  2/10 &  28\\
\addlinespace[2pt]
{w/o Keyframe Thinking} & 14/20 & 15/20 & 3/10 & 64 \\
\midrule
\textbf{Ours} &  \textbf{17/20} & \textbf{18/20} & \textbf{6/10} &  \textbf{82}\\
\bottomrule
\end{tabular}
\vspace{4pt}
\vspace{-0.5cm}
\end{table}

\begin{table}[t]
\centering
\caption{Results of In-the-wild task.}
\label{tab:in_the_wild}
\vspace{-2mm}
\setlength{\tabcolsep}{4pt}

\begin{tabular}{l|c c c|c}
\toprule
\textbf{Scenario} &\textbf{1 Object} & \textbf{2 Objects} & \textbf{3 Objects} & \textbf{SuccR} (\%) \\
\midrule
Open Ground & 5/5 & 4/5 & 4/5 & 86.7 \\
\addlinespace[2pt]
Shaded Path & 5/5 & 3/5 & 2/5 &  66.7\\
\addlinespace[2pt]Grassy Ground & 4/5 & 3/5 & 3/5 & 66.7\\
\addlinespace[2pt]
Stepped Ground & 4/5 & 3/5 & 2/5 & 60.0\\
\bottomrule
\end{tabular}

\vspace{2pt}
\vspace{-0.6cm}
\end{table}

\begin{table}[t]
\centering
\caption{Robustness to External Interference.}
\label{tab:recovery}
\vspace{-2mm}

\begin{tabular}{l|c c c|c}
    \toprule
    \multirow{2}{*}{\textbf{{Method}}} &
    \textbf{Table} &
    \textbf{Ground} &
    \textbf{Chest} & 
    \multirow{2}{*}{\textbf{SuccR(\%)}} \\
    &
    \textbf{Grasping} &
    \textbf{Grasping} &
    \textbf{Manip.} & 
    \\    
    \midrule

{ACT} & 1/5 & 2/5 & 1/5 & 26.7\\
\addlinespace[2pt]
${\pi_{0.5}}$ & 2/5 & 2/5 & 1/5 &  33.3\\
\addlinespace[2pt]
{plan-then-act} & 0/5 & 0/5 & 0/5 & 0.0 \\
\midrule
\textbf{Ours} & 4/5 & 5/5 & 2/5 & 73.3 \\
\bottomrule
\end{tabular}

\vspace{4pt}
\vspace{-0.5cm}
\end{table}

\begin{figure}[t]
    \centering
    \includegraphics[width=0.49\textwidth]{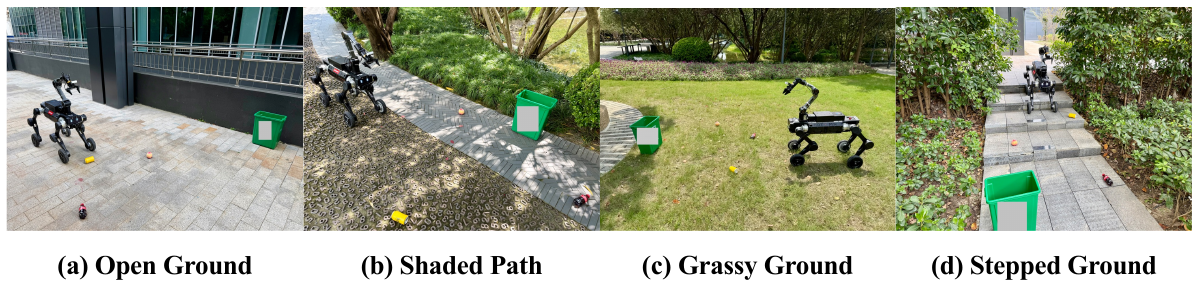}
    \caption{\textbf{Ground Grasping task in outdoor scenes.}}
    \label{fig:in-the-wild}
    \vspace{-8mm}
\end{figure}

We also conduct several experiments on three indoor tasks to verify the effectiveness of each design of our model and the results are shown in Table \ref{tab:ablation}. We can see that locomotive tools for whole-body control play a important role in loco-manipulation tasks to handle complex loco-manipulation tasks at varying heights. Besides, removing keyframe thinking mechanism resulting a decline in three tasks. We think the reason is that keyframe information provides valuable cues for the agent to identify the causes of execution failures and take appropriate recovery actions.

\subsection{In-The-Wild Task}

To further evaluate the robustness and generalization capability of our method, we conduct several in-the-wild long-horizon loco-manipulation experiments. Four different outdoor scenes are selected to perform the \textit{Ground Grasping} task as seen in Fig. \ref{fig:in-the-wild}. 
We set three tasks involving sequential grasp-and-place operations with one, two, and three objects, respectively. 
The results are shown in Table \ref{tab:in_the_wild}, demonstrating strong robustness and generalization capability of our method across diverse terrains.

\vspace{-0.1cm}
\subsection{Robustness to External Interference}
We also conduct experiments on three indoor tasks to evaluate the robot’s robustness to external interference. We manually reset the task environment after each successful trial, such as replacing the grasped objects or reclosing the chest after successful opening. The results are shown in Table \ref{tab:recovery}. The Plan-then-Act method fails to recover because its task plan is fixed after generation and cannot be updated in response to execution failures. We also observe that although ACT and $\pi_{0.5}$ tend to resume task execution after environment resets, such as moving the gripper close to the object, they still fail to complete the tasks successfully, demonstrating their sensitivity to external interference. Our method achieves the best performance, which can be attributed to the design of the Error Recovery skill and the Keyframe Thinking mechanism. This enables the agent to accurately diagnose failure causes and promptly adapt its behavior to recover from execution failures.

\section{CONCLUSION}

In this work, we present \textbf{ThinkLocoManip}, an agentic harness for complex, long-horizon legged loco-manipulation in large-scale real-world environments. Rather than directly generating actions or committing to a fixed plan before execution, ThinkLocoManip adopts an \textit{interleaved reasoning-and-acting} paradigm, allowing the agent to continuously reason about the current state, invoke appropriate tools, observe execution outcomes, and adapt its strategy accordingly. Building on this paradigm, we organize specialized primitive tools for perception, navigation, manipulation, and locomotion, and introduce \textbf{ThinkLocoManip-Skill} to structure tool usage and cross-domain coordination. We further develop \textbf{Keyframe Thinking} to preserve critical execution context and support efficient error recovery during long-horizon interactions. Extensive real-world experiments demonstrate that ThinkLocoManip substantially outperforms representative direct action generation and plan-then-act methods, achieving an overall success rate of \textbf{77.0\%}, compared with \textbf{37.1\%} for the strongest baseline. Additional ablations, external-interference experiments, and in-the-wild evaluations further validate the importance of whole-body control and Keyframe Thinking, as well as the robustness and generalization capability of the overall framework. We hope ThinkLocoManip provides a step toward more general-purpose robotic agents that can flexibly compose heterogeneous capabilities, adapt to intermediate execution outcomes, and robustly accomplish long-horizon loco-manipulation tasks in complex real-world environments.

\vspace{-2mm}


\bibliographystyle{IEEEtran}
\bibliography{cite}

\end{document}